\documentclass[3p, review]{elsarticle}

\usepackage{lineno,hyperref}
\usepackage{booktabs}
\modulolinenumbers[5]

\usepackage[figuresright]{rotating}
\usepackage{amsopn}
\usepackage{amssymb}
\usepackage{amsmath}
\usepackage{subfig}
\usepackage{array}
\usepackage{tabularx}
\usepackage{multirow}
\usepackage{booktabs}
\usepackage{graphicx}
\usepackage{mathtools}
\usepackage{algorithmic}
\usepackage{algorithm}
\usepackage{listings}
\usepackage{setspace}
\usepackage{url}
\usepackage{float}
\usepackage{epsf} 
\usepackage{tikz}
\usepackage{amsthm}
\usepackage{color}
\usepackage{lineno}
\usepackage{etoolbox} 

\usepackage{multirow}
\usepackage{graphicx}
\usepackage[normalem]{ulem}
\useunder{\uline}{\ul}{}

\theoremstyle{definition}
\usepackage{url}

\journal{Engineering Applications of Artificial Intelligence}

\makeatletter
\patchcmd{\ps@pprintTitle}
  {Preprint submitted to}
  {Accepted by}
  {}{}
\makeatother

\begin{document}

\begin{frontmatter}

\title{DFNet: Discriminative feature extraction and integration network for salient
object detection}


\author[a]{Mehrdad Noori\fnref{equal}}
\ead{me.noori.1994@gmail.com}
\fntext[equal]{Both the authors contributed equally.}

\author[a]{Sina Mohammadi\fnref{equal}}
\ead{sina.mhm93@gmail.com}

\author[a]{Sina Ghofrani Majelan}
\ead{sghofrani1@gmail.com}

\author[a]{Ali Bahri}
\ead{alibahri.72.dl.k@gmail.com}

\author[b]{{Mohammad Havaei}\corref{Corresponding}}
\ead{mohammad@imagia.com}
\cortext[Corresponding]{Corresponding author.}
    
\address[a]{School of Electrical Engineering\unskip, 
    Iran University of science and Technology\unskip, Tehran, Iran}
  	
\address[b]{
    Imagia Inc.\unskip, Montreal, Canada}

\begin{abstract}

Despite the powerful feature extraction capability of Convolutional Neural Networks, there are still some challenges in saliency detection. In this paper, we focus on two aspects of challenges: i) Since salient objects appear in various sizes, using single-scale convolution would not capture the right size. Moreover, using multi-scale convolutions without considering their importance may confuse the model. ii) Employing multi-level features helps the model use both local and global context. However, treating all features equally results in information redundancy. Therefore, there needs to be a mechanism to intelligently select which features in different levels are useful. To address the first challenge, we propose a Multi-scale Attention Guided Module. This module not only extracts multi-scale features effectively but also gives more attention to more discriminative feature maps corresponding to the scale of the salient object. To address the second challenge, we propose an Attention-based Multi-level Integrator Module to give the model the ability to assign different weights to multi-level feature maps. Furthermore, our Sharpening Loss function guides our network to output saliency maps with higher certainty and less blurry salient objects, and it has far better performance than the Cross-entropy loss. For the first time, we adopt four different backbones to show the generalization of our method. Experiments on five challenging datasets prove that our method achieves the state-of-the-art performance. Our approach is fast as well and can run at a real-time speed.

\end{abstract}


\end{frontmatter}

\doublespacing

\section{Introduction}

Saliency detection in computer vision is the process to determine the most prominent and conspicuous parts of an image. Selective attention is embedded in our cognitive system and a lot of the tasks we do in every day life
depend on it. Saliency detection has applications in a variety of supervised and unsupervised
tasks~\cite{frintrop2008attentional,walther2006modeling,zdziarski2012feature,bi2014person,hong2015online,mahadevan2009saliency,ren2013region}. For example, salient object detection can provide informative prior knowledge
to objectness detection. The extracted bounding box locations which are more prominent and salient in an image would be more likely to contain the objects of interest~\cite{han2018advanced}. Due to this fact, some
objectness detection methods use saliency cues to detect objects of interest~\cite{alexe2012measuring,erhan2014scalable}.

The traditional computer vision approach to saliency detection is to identify parts of the image that have different contextual information with respect to their surroundings. To identify salient parts of an image, we would require both local and global contextual information. While local contextual features can help to reconstruct the object boundaries, global contextual features are beneficial for getting an abstract description of the salient object.

\begin{figure}[t]
\begin{center}
\includegraphics[width=0.75\textwidth]{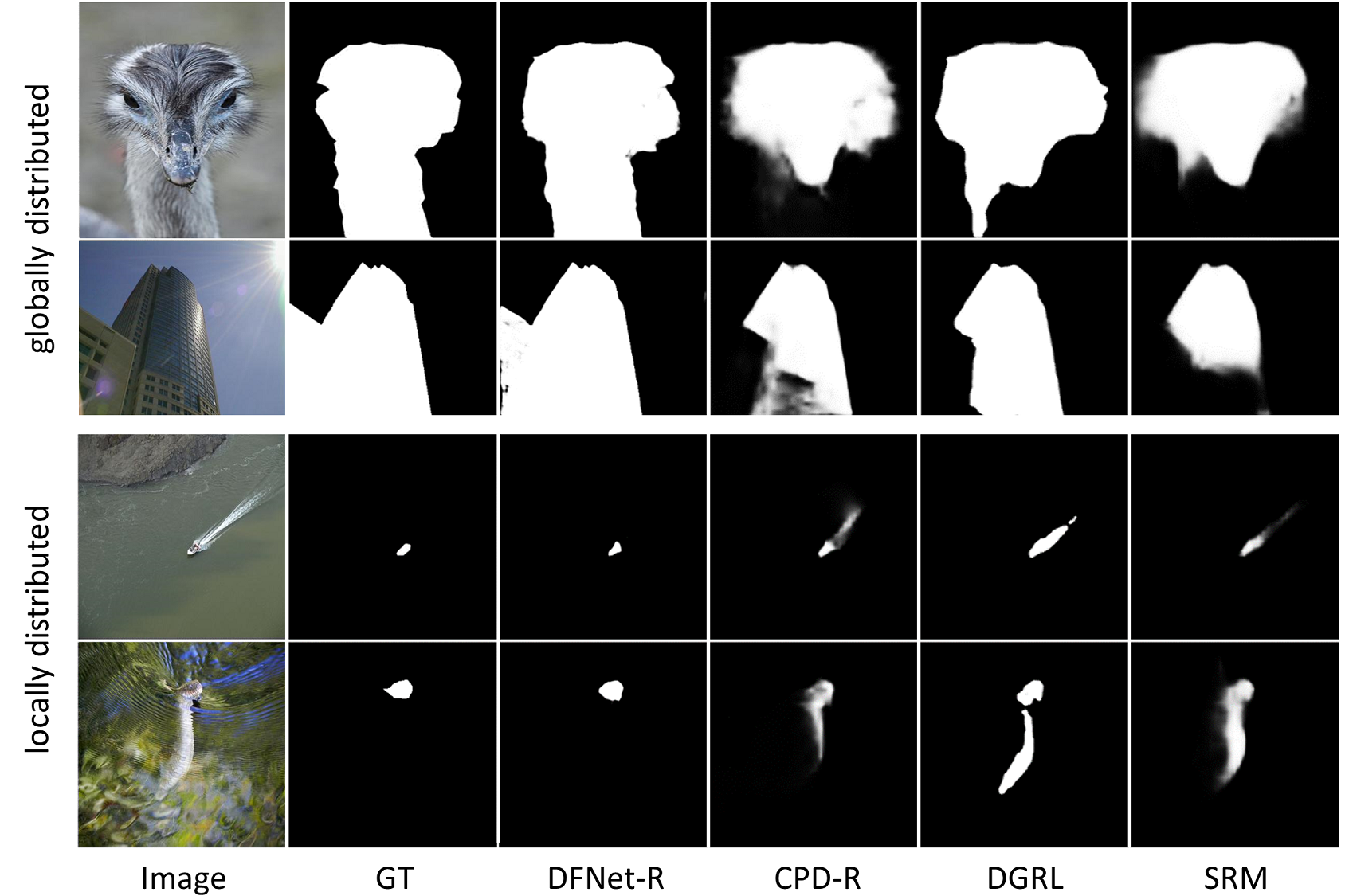}
\end{center}
   \caption{Two challenging scenarios of saliency detection. In the first scenario, the salient object is globally distributed. In the second one, the salient object is locally distributed. While both scenarios have caused
confusion for three recent methods (CPD-R~\cite{wu2019cascaded}, DGRL~\cite{wang2018detect}, and SRM~\cite{wang2017stagewise}), our method (denoted as DFNet-R) is capable of handling these challenging scenarios by
benefiting from the MAG Modules adopted in our model.}
\label{fig:first_page}
\end{figure}

With the ability of deep learning models in extracting high-level features, some early papers used these models to extract features from candidate image regions at different resolutions to extract local and global
representations of the salient objects~\cite{lee2016deep,zhao2015saliency,li2015visual,wang2015deep}. Despite their success, due to the use of dense layers, these methods were not very efficient. However, deep neural
networks inherently extract increasingly complex features from low-level  to high-level and so in recent years, many papers have tried to use features from different levels of abstraction to incorporate low-level features
and the  more global high-level features. Figuring out how to combine the two information is still an open question. While the conventional way is to concatenate the low-level features with high-level features, and thus
treating all feature maps equally, we propose to use an adaptive concatenation functionality where conditioned on the input, the model re-weights the concatenating features. To achieve this purpose, we introduce the
Attention-based Multi-level Integrator (AMI) Module, which first weights the concatenated multi-level features by using a Channel Attention (CA) Block, and then it refines the resulted features by using a convolutional
layer. Note that the CA block is similar to the recently introduced squeeze and excitation (SE) networks~\cite{hu2018squeeze}.

To be able to capture salient object in different sizes, Inception like~\cite{szegedy2015going} modules can be used to extract features at different receptive fields~\cite{szegedy2015going}.  Previous works concatenate
features from different scales which means assigning equal importance to all scales. While such functionality is desirable for applications like image segmentation, for saliency detection we usually consider a single scale
as the salient object. The ability to capture the right size for the salient object can be achieved by assigning dynamic weights to the output feature maps of the Inception module, where conditioned on the input image, the model gives different importance to different scales.
To achieve this functionality, we introduce a Multi-scale Attention Guided (MAG) Module. By using a novel design, this module first extracts multi-scale features effectively, and then it adaptively gives different
importance to different scales by adopting the Channel Attention Block. In Figure \ref{fig:first_page}, two challenging scenarios of saliency detection are shown. In the first scenario, the salient object is globally distributed
over the image. In the second one, the salient object is locally distributed. As seen from Figure \ref{fig:first_page}, our method (denoted as DFNet-R) is able to handle these challenging scenarios, unlike three recent methods.
This functionality is achieved by using MAG Modules in our framework. Therefore, if the salient information is spread globally, the model will give more attention to feature maps from larger kernels. While, if the salient
information is spread locally the model will emphasize feature maps of smaller kernels. 

In this paper, we propose a Discriminative Feature Extraction and Integration Network, which we refer to as DFNet, consisting of two parts; $(i)$  the Feature Extraction Network and $(ii)$ the Feature Integration Network. In the Feature Extraction Network, by adopting the MAG Modules, we extract dynamically weighted multi-scale features from a pre-trained network at various levels of abstraction. These features are then combined together in the Feature Integration Network by employing the AMI Modules. It is interesting to note that while using a single pre-trained network as the backbone is a common practice in saliency detection, for the first time in the literature, we use four different backbones in our framework to prove the robustness and generalization capability of our method. Furthermore, while the Cross-entropy loss is widely used in the literature, we discover that using this loss function leads to blurry predictions, as we show in the ablation study section. To boost the certainty of our proposed model, we design a Sharpening Loss function, which forces our network to generate sharper predictions. Through experiments, we show that our designed loss outperforms the Cross-entropy loss by a large margin. It is worth mentioning that saliency detection is a pre-processing step for various computer vision tasks. Since our method can run at a real-time speed, it can be practically adopted as a pre-processing step.

In summary, the contributions of this paper are four fold:
\begin{itemize}
    \item We propose a Multi-scale Attention Guided (MAG) Module to give our model the ability to capture the right size for the salient object. This module not only can effectively extract multi-scale features by adopting a novel implementation, but also can give more attention to more discriminative feature maps and attenuate feature maps not corresponding to the scale of the salient object in the input image. 
    \item We propose an Attention-based Multi-level Integrator (AMI) Module to equip our model with the power of assigning different weights to multi-level feature maps, which would let the beneficial features participate more in producing the saliency map.
    \item We design a loss function which guides our network to output saliency maps with higher certainty. This loss function results in sharper foreground objects and less blurry predictions. Moreover, as we show in ablation study section, our designed loss outperforms the widely-used Cross-entropy by a significant margin.
    \item Our method achieves the state-of-the-art on five challenging datasets. Furthermore, for the first time in the saliency detection literature, we adopt four different backbones in our framework to prove  the robustness and generalization capability of our method.

\end{itemize}

\section{Related work}
Due to the importance of salient object detection and its vast application in various fields, a lot of works have focused on salient object detection over the past decade. Early works for solving this problem was mainly
based on low-level cues such as color, texture,
contrast~\cite{perazzi2012saliency,cheng2014global,jiang2013salient,li2014secrets,borji2015salienta,borji2015salientb,yan2013hierarchical,yang2013saliency,achanta2009frequency}.
 More recently with the success of neural networks in learning high-level task-specific features, a lot of effort has been made to develop models to extract features for detecting salient regions. For example, 
 Han et al. \cite{han2017unified} use a convolutional neural network for initial feature extraction. Then they proposed a metric learning-based co-saliency detection method to simultaneously learn discriminative feature
representations and co-salient object detector. 

The methods based on neural networks can be divided into two main subcategories; patch-based models and models based on fully convolutional neural networks (FCN). 
In patch-based models, a patch is extracted around each pixel. The neural network would then assign a saliency score for the center pixel of every patch.  Li and Yu \cite{li2015visual} use patches at different sizes to
extract multi-scale features for every pixel. These features were concatenated and fed into a classifier to determine the saliency score assigned to that pixel. Wang et al. \cite{wang2015deep} use a two path model to
refine features from patches containing local context with features from object proposals which contain more global context.

Fully convolutional networks have the ability to ingest the whole image, do not require the input to have a fixed size, and thus provide for more flexibility and efficiency compared to patch-based 
models. Xi et al. \cite{xi2019salient} propose an efficient end-to-end FCN architecture based on saliency regression network, which directly outputs a dense full-resolution saliency map for a given 
input image. Wang et al. \cite{wang2016saliency} use recurrent neural networks to iteratively refine saliency maps extracted via an FCN. 
In another work, Wang et al. \cite{wang2017stagewise} use multiple stages of FCN to refine the
saliency maps from previous stages.
A number of methods have attempted to combine low-level features with high-level features using skip connections. Liu and Han \cite{liu2016dhsnet} use long skip connections in a U-Net like architecture to concatenate
high-level features and low-level features for saliency detection. 
Zhang et al. \cite{zhang2018bi} also concatenate low-level features with high-level features in bi-directional way. Zhang et al. \cite{zhang2017amulet}
extracts features from different layers and concatenates them before passing them to a refinement module. While the idea of using both low-level and high-level features for saliency detection is not new, what seems to be
lacking is a mechanism which allows the model to intelligently select which level of abstraction it needs given the input image. In this work, we propose a simple yet effective architecture to achieve this.

Using kernels at varying sizes in a manner of Inception module  is a way to capture features at multiple scales. They have shown to be successful in applications like semantic segmentation where we expect to capture
objects at multiple scales~\cite{kim2017new}. Inception module has also been used in some saliency detection methods~\cite{chen2017look,zhang2018bi}. In all these models, feature maps are extracted by applying kernels of
multiple sizes. These feature maps are concatenated before being passed to the next layer. In this paper, we propose an architectural design for Inception module where for every input image the model assigns different
weights (i.e.~importance) to different feature maps resulted from kernels of multiple sizes. Therefore, the model has the flexibility to focus its attention to more discriminative feature maps and discard the information
clutter it may receive from other feature maps.

\section{The Proposed Method}
In this section, we explain our proposed method for saliency detection task. We firstly describe the two parts of our DFNet, namely the Feature Extraction Network and the Feature Integration Network, in Section \ref{ch:feature-extraction-network} and \ref{ch:feature-integration-network}. Then, we proceed with explaining the objective function we used to learn sharper salient objects in Section \ref{ch:learning_sharper}. The architecture of the proposed DFNet is illustrated in Figure \ref{fig:Main}, and an overview of the different components of DFNet is depicted in Figure \ref{fig:all_modules}.

\subsection{Feature Extraction Network}\label{ch:feature-extraction-network}
The main functionality of the Feature Extraction Network is to extract representative local and global features at multiple scales in order to be used by the \textit{Feature Integration Network}. This network is composed of two main parts; \textit{Backbone} and \textit{Multi-scale Attention Guided Module (MAG Module)}. We explain each part in detail.

\subsubsection{Backbone}
In the saliency object detection task, an ImageNet~\cite{russakovsky2015imagenet} pre-trained model is often used as the backbone to extract a hierarchy of increasingly complex features at different levels of abstraction.
One of the advantages of our approach is that it is very flexible and can be used with any \textit{backbone} without the need to change the architecture of the rest of the model. In the DFNet framework, we examine
VGG-16~\cite{simonyan2014very}, ResNet50~\cite{he2016deep}, NASNet-Mobile~\cite{zoph2018learning}, and NASNet-large~\cite{zoph2018learning} as the backbone, which are denoted as DFNet-V, DFNet-R, DFNet-M, and DFNet-L,
respectively. The backbones are pre-trained to extract features for image classification. However, since we are dealing with assigning per-pixel saliency score, we make modifications to these models to fit the need of
saliency detection task. 
To this end, we remove all the dense layers in the backbones. Since each backbone has a different architecture, there needs to be a selection process in terms of which layers to select the feature maps from. In what follows, we explain this selection process for every backbone:

VGG-16 has 5 max pooling layers. We remove the last pooling layer to retain a better spatial representation of the input.  We utilize feature maps of the last 3 stages from the VGG-16: conv3-3 (256 feature maps), conv4-3 (512 feature maps), and conv5-3 (512 feature maps).

\begin{figure}[t]
\begin{center}
\includegraphics[width=1\textwidth]{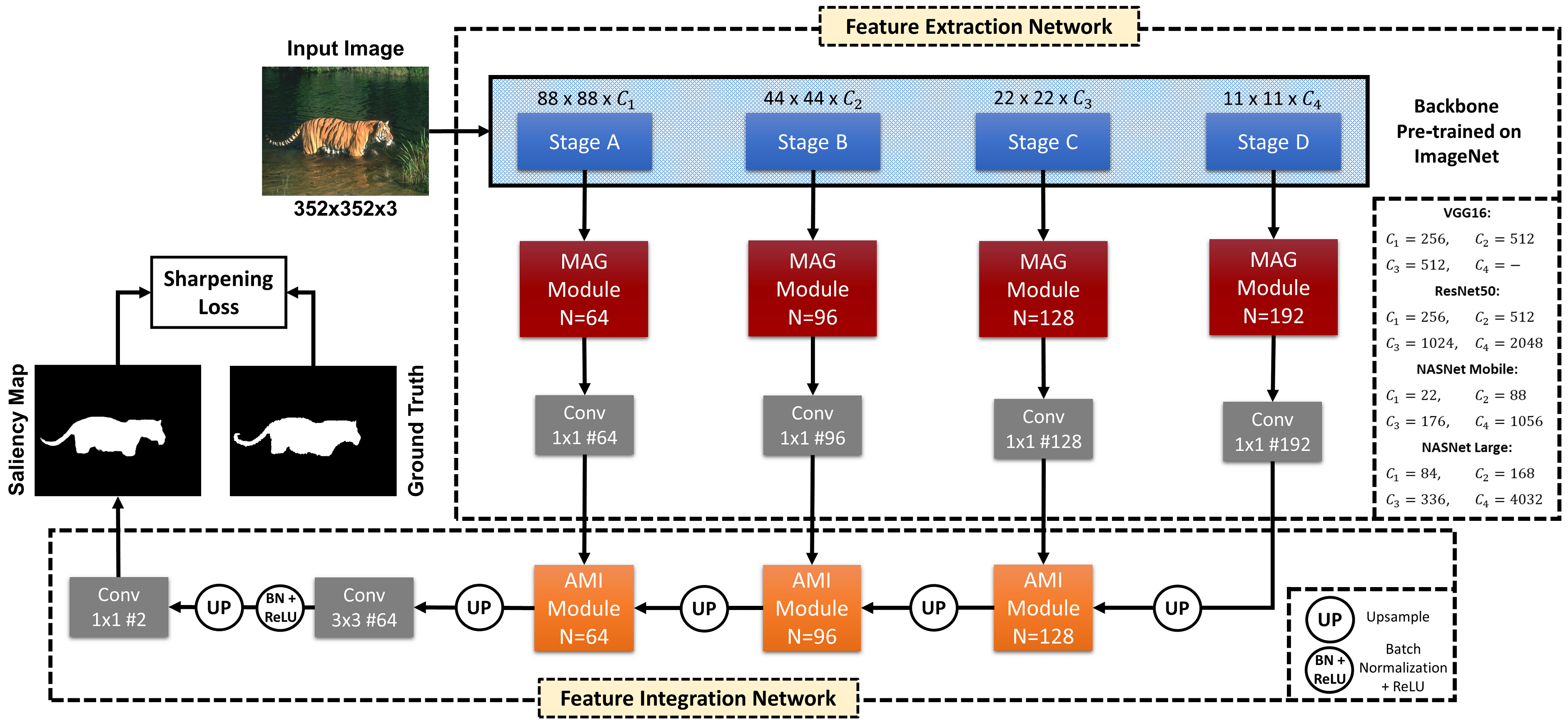}
\end{center}
   \caption{Architecture of the proposed DFNet. Our model is composed of two networks: (i) Feature Extraction Network, which extracts representative features at multiple scales. (ii) Feature Integration Network, which fuses multi-level features effectively. Note that this figure represents the model
architecture when ResNet50, NASNet-Mobile, and NASNet-Large are used as
the backbone. However, when VGG-16 is adopted as the backbone, there
are only three stages instead of four. We remedy this problem by
rewiring the branches connected to stage B, C, D to A, B, C when VGG-16
is used. So for example, for stage A in VGG-16, we have MAG Module and
AMI Module with $N=96$.}
\label{fig:Main}
\end{figure}

\begin{figure}[H]
\begin{center}
\includegraphics[width=1\textwidth]{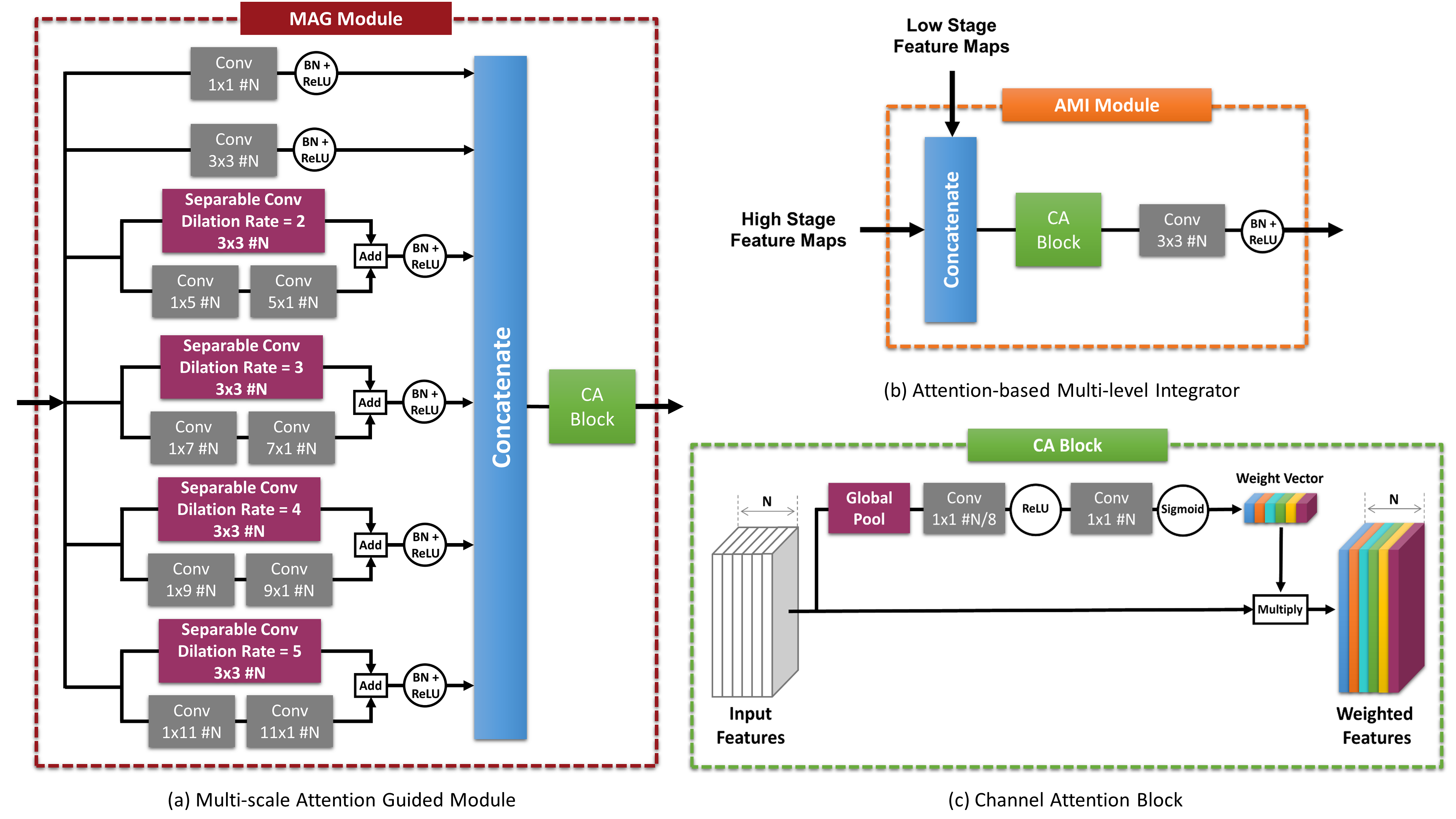}
\end{center}
   \caption{An overview of the different components of our proposed DFNet. (a) Multi-scale Attention Guided Module. This module performs convolutions with kernels of multiple sizes. Then, after concatenation, we use the Channel Attention Block to weight the multi-scale features. (b) Attention-based Multi-level Integrator Module. This module first concatenates high stage features with low stage features. Then the Channel Attention Block is used to assign different weights to multi-level features. Finally, a $3\times3$ convolutional layer is used to refine the features. (c) Channel Attention Block. This block computes a weight vector to re-weight the input feature maps. Note that in all figures, the `\#' symbol denotes the number of layer filters.}
\label{fig:all_modules}
\end{figure}

ResNet50, which consists of 5 residual convolution blocks, has 5 stages with different spatial resolutions. We use feature maps of the last 4 stages,  namely conv2-x (256 feature maps), conv3-x (512 feature maps), conv4-x (1024 feature maps), and conv5-x (2048 feature maps).

NASNet has a very complicated architecture, and thus mentioning the layers name from which we extract features, needs a detailed illustration of the NASNet architecture. Therefore, we encourage the readers to refer to the publicly available code for more details on which layers we used. In this section, we just provide the number of feature maps of each stage. In the case of NASNet-Mobile and NASNet-Large, we use four stages. In NASNet-Mobile, these stages contain $22$, $88$, $176$, and $1056$ feature maps, respectively. In NASNet-Large, the number of feature maps increase to $84$, $168$, $336$, and $4032$, respectively.

To make things clear, considering the backbone with feature map sizes of $\frac{W}{2^n} \times \frac{H}{2^n}$, we utilize feature map sizes with $n=2$, $3$, $4$ for VGG-16 and  feature map sizes with $n=2$, $3$, $4$, $5$ for ResNet50, NASNet-Mobile, and NASNet-Large. The extracted feature maps on each level (i.e.~stage) are passed through the \textit{MAG Module}, which is explained next.

\subsubsection{Multi-scale Attention Guided Module}
It is evident that large kernels are suitable to capture the large objects, and small kernels are appropriate to capture the small ones. Due to the size variability of salient objects, it is not the best  approach to
employ simple, fixed-size kernels. Therefore, to capture objects of different scales at the same time, we adopt kernels in various sizes in an Inception~\cite{szegedy2015going} like fashion. More specifically, we perform
convolutions with $1\times1$, $3\times3$, $5\times5$, $7\times7$, $9\times9$, and $11\times11$ kernels. Then, the resulting feature maps are stacked to form multi-scale features. The idea of extracting
multi-scale features via inception modules has been previously explored. The difference between our method and the existing ones is that we also employ an attention mechanism to weight feature maps of various scales, and
thus the model learns to give more attention to the right size and attenuate feature maps not corresponding to the scale of the salient object in the input image. In other words, if the salient information is spread
globally, the model will put more weight on feature maps from larger kernels and if the salient information is spread locally the model will emphasize feature maps of smaller kernels. From our point of view, giving the
model this ability and flexibility is the key factor to enhance the overall performance and avoid confusion for the model, which was the missing link in the previous works. Additionally, the implementation of this module,
which is described next, is novel and efficient in term of memory.

\paragraph*{Implementation of MAG Module}
 Convolutions with large kernel sizes such as $5\times5$ and higher are computationally very expensive. We adopt two solutions to mitigate this problem: ($i$) We can factorize an $n\times n$ kernel to a combination of $1\times n$ and $n\times 1$ kernels, ($ii$) An $n\times n$ kernel with dilation rate of $r$ will have the same receptive field as a kernel of size $(n+(r-1)\times 2) \times (n+(r-1) \times 2)$. Our MAG Module employs a combination of these two approaches to implement an $n\times n$ kernel.
To weight the multi-scale features, we use the CA Block which is illustrated in Figure \ref{fig:all_modules}(c). This design computes a weight vector to re-weight input feature maps. The implementation of MAG Module is shown in Figure \ref{fig:all_modules}(a). MAG Module is used in every branch of the Feature Extraction Network as shown in Figure \ref{fig:Main}. In every branch after MAG Module, we use a $1\times1$ convolutional layer to combine the feature maps and reduce the number of them.

\subsection{Feature Integration Network}\label{ch:feature-integration-network}
By employing MAG Modules, effective multi-scale contextual information at different levels is captured, as illustrated in Figure \ref{fig:Main}. In order to effectively integrate the multi-level features, we introduce the Feature Integration Network. As described in Section \ref{ch:feature-extraction-network}, the \textit{Feature Extraction Network} extracts features at four stages (three as for VGG-16 backbone). These stages contain diverse recognition information. At lower stages, the network captures such local structures as textures and edges. However, it fails to recognize global dependencies due to its small field of view. On the other hand, at higher stages, the model captures semantics and the global context of the image due to its large effective receptive field.  However, at this stage, the information is very coarse and lacks the local consistency we observed in the lower stages. Since both type of features are necessary for saliency detection, to take advantage of both worlds, we introduce Attention-based Multi-level Integrator Module (AMI Module), where the semantic information in high-level features and the spatial details in low-level features are effectively combined. In the AMI Module, features from different stages of the Feature Extraction Network are concatenated, followed by the CA Block to weight each feature map. The nature of low-level features and high-level features is very different, and thus combining them uniformly through concatenation with uniform weights may not be the best procedure. 
Using the CA Block in this module will give the model the ability and flexibility to assign different weights to semantic information and spatial details. After the CA Block, a $3\times3$ convolutional layer is used to refine the features. The architecture of the AMI Module is shown in Figure \ref{fig:all_modules}(b). As illustrated in Figure \ref{fig:Main}, in the Feature Integration Network, the saliency map is generated by using AMI Modules, a series of upsampling layers, and convolution layers. By using this structure, the feature maps from different levels can collaborate to generate a more accurate prediction.

\subsection{Learning Sharper Salient Objects}\label{ch:learning_sharper}
The Cross-entropy loss is widely-used for learning the salient objects. We discover that using this loss function in the salient object detection task leads to blurry and uncertain predictions. To learn sharper salient objects, we design a loss function, which we refer to as the Sharpening Loss, defined as:
\begin{equation}
L_S=L_F+\lambda\cdot L_{MAE}
\end{equation}
 where $\lambda$ is used to balance the F-measure loss $L_F$ and the MAE loss $L_{MAE}$. $\lambda$ is empirically set to 1.75.  We denote the training images as $I = \{I_m,~m=1, \ldots ,M\}$. $S_m$ is the saliency map, and $G_m$ is the ground truth map for $m$-th training image. $L_F$  is computed as:
 \begin{equation}
L_F=1-\frac{(1+\beta^2)\cdot \sum_{m=1}^M \frac{P{(S_m,G_m)}}{M}\cdot \sum_{m=1}^M \frac{R{(S_m,G_m)}}{M}}{\beta^2\cdot \sum_{m=1}^M \frac{P{(S_m,G_m)}}{M}+\sum_{m=1}^M \frac{R{(S_m,G_m)}}{M}+\epsilon}
\end{equation}
 where $\beta^2$ is set to 0.3 as suggested in~\cite{yang2013saliency}, and $\epsilon$ is a regularization constant. Since higher values of F-measure are better, subtraction of it from $1$ is used for
minimizing. $P{(S,G)}$ and $R{(S,G)}$ are calculated similar to Precision and Recall: 
\begin{equation}
P{(S,G)}=\frac{\sum_i s_i\cdot g_i}{\sum_is_i+\epsilon}
\end{equation}
\begin{equation}
R{(S,G)}=\frac{\sum_i s_i\cdot g_i}{\sum_ig_i+\epsilon}
\end{equation}
where $s_i \in S$ and $g_i \in G$. $L_{MAE}$ is used to calculate the discrepancy between the predicted saliency map $S$ and the ground truth map $G$: 
\begin{equation}
L_{MAE}=\frac{1}{M}\sum_{m=1}^{M}MAE(S_m, G_m)
\end{equation}
 where $MAE(S, G)$ is computed as:\begin{equation}
MAE(S, G) = \frac{1}{N}\sum_i \mid s_i - g_i \mid
\end{equation}
 where $N$ is the total number of pixels.

We compare the designed loss function with the Cross-entropy loss in the ablation study section, and we will show that the Sharpening Loss gives better results and sharper salient objects compared to the Cross-entropy loss.

\section{Experiments}
\subsection{Datasets}
 We evaluate the proposed method on five public saliency detection datasets which are human-labeled with pixel-level ground truth. DUTS~\cite{wang2017learning} is a large scale salient object detection benchmark dataset
comprised of 10553 images for training and 5019 images for testing. Most of the images contain complex and challenging scenarios. ECSSD~\cite{yang2013saliency} contains 1000 images with complex scenes and objects of
different sizes. HKU~\cite{li2015visual} consists of 4447 images. Most images in this dataset include multiple disconnected salient objects or objects touching the image boundary with low color contrast.
PASCAL-S~\cite{li2014secrets} contains 850 natural images generated from the PASCAL VOC dataset~\cite{everingham2010pascal} which has complex images due to cluttered backgrounds and multiple objects.
DUT-OMRON~\cite{yang2013saliency} includes 5168 complex and challenging images with high content variety. Images in this dataset have one or more salient objects and complex background.

\subsection{Evaluation metrics}\label{sec:evaluation_metrics}

We utilize Precision-Recall (PR) curve, F-measure curve, Average F-measure (avgF) score, Weighted F-measure (wF) score, Maximum F-measure (maxF) score, and Mean Absolute Error (MAE) score as our evaluation metrics.

Precision is defined as the fraction of salient pixels labeled correctly in the predicted saliency maps, and Recall is the fraction of salient pixels labeled correctly in the ground truth. To calculate Precision and Recall, predicted saliency maps are binarized by thresholding, and compared with the ground truth. The F-measure score is a metric for overall performance which considers both Precision and Recall:
\begin{equation}
F_\beta=\frac{(1+\beta^2)\cdot Precision\cdot Recall}{\beta^2\cdot Precision + Recall}
\label{eqn:F_metric}
\end{equation}
 where  $\beta^2$ is set to 0.3, as suggested in~\cite{yang2013saliency} to emphasize the precision.

To plot the PR curve, binarization of the saliency maps is done under different thresholds. Thus, a series of binary maps are obtained. Then from these binary maps, Precision, Recall, and F-measure values can be calculated. The obtained values of (Precision, Recall) pairs and (F-measure, threshold) pairs are employed to plot the PR curve and the F-measure curve. 

Average F-measure score is computed by using the thresholding method suggested in~\cite{achanta2009frequency}. This threshold, which is twice the mean saliency value of each saliency map, is used to generate binary maps
for computing the Average F-measure. Weighted F-measure score is calculated by introducing a weighted Precision to measure the exactness and a weighted Recall to measure the completeness (refer to~\citealp{margolin2014evaluate} for
more details). Maximum F-measure score is reported as the maximum value in the F-measure curve. Furthermore, we report the MAE score which is calculated as the average pixel-wise absolute difference between the binary
ground truth $G$ and the predicted saliency map $S$:
\begin{equation}
MAE=\frac{1}{W\times H}\sum_{x=1}^W \sum_{y=1}^H \mid S_{(x,y)}- G_{(x,y)} \mid
\end{equation}
 where $W$ and $H$ denote width and height of $G$.

\subsection{Implementation details}

    DFNet is developed in Keras~\cite{chollet2015keras} using TensorFlow~\cite{tensorflow2015-whitepaper} backend. An NVIDIA 1080 Ti GPU is used for training and testing. The training set of DUTS dataset is utilized to train
our network for salient object detection. In our experiments, all input images are resized to $352\times352$ pixels for training and testing. To reduce overfitting, two kinds of data augmentations are employed at
random: horizontal flipping and rotation (range of 0--12 degrees). We do not use validation set and train the model until its training loss converges. We use the stochastic gradient descent with a momentum coefficient 0.9,
and a base learning rate of $8e$ $-3$. If the training loss does not decrease for ten epochs, the learning rate is divided by $10$. The code and the saliency maps of our method can be found at {{\color[HTML]{3166FF}{\href{https://github.com/Sina-Mohammadi/DFNet}{https://github.com/Sina-Mohammadi/DFNet}}}

\subsection{Comparison with the State-of-the-Art}

We compare the proposed saliency detection method against previous 18 state-of-the-art methods, namely, MDF~\cite{li2015visual}, RFCN~\cite{wang2016saliency}, DHS~\cite{liu2016dhsnet}, UCF~\cite{zhang2017learning},
Amulet~\cite{zhang2017amulet}, NLDF~\cite{luo2017non}, DSS~\cite{hou2017deeply}, RAS~\cite{chen2018reverse}, BMPM~\cite{zhang2018bi}, PAGR~\cite{zhang2018progressive}, PiCANet~\cite{liu2018picanet},
SRM~\cite{wang2017stagewise}, DGRL~\cite{wang2018detect}, MLMS~\cite{wu2019mutual}, AFNet~\cite{feng2019attentive}, CapSal~\cite{zhang2019capsal}, BASNet~\cite{qin2019basnet}, and CPD~\cite{wu2019cascaded}. We
perform comparisons on five challenging datasets. For fair comparison, we evaluate every method by using the saliency maps provided by the authors.

For quantitative comparison, we compare our method with previous state-of-the-art methods in terms of the PR curve, F-measure curve, avgF, wF, maxF,  and MAE. The PR curves and F-measure curves on five datasets are shown
in Figure \ref{fig:PRcurve} and Figure \ref{fig:Fcurve}, respectively. We can observe that our proposed model performs favorably against other methods in all cases. Especially, it is evident that our DFNet-L performs better than all
other methods by a relatively large margin. Additionally, the avgF scores, wF scores, maxF scores,  MAE scores, and the total number of parameters of different methods are provided in Table \ref{tab:comparison}. As seen in
the table, considering all four backbones, our method outperforms other state-of-the-art methods in most cases. Comparing Average F-measure scores (avgF in Table \ref{tab:comparison}), our DFNet-L improves the value by
$7.4\%$, $2.2\%$, $4.4\%$, $4.9\%$, $3.1\%$ on DUTS-TE, ECSSD, DUT-O, PASCAL-S, HKU-IS, respectively. In addition, our DFNet-L lowers the MAE scores by $23.2\%$, $24.3\%$, $7.1\%$,
$23.9\%$, $12.5\%$ on DUTS-TE, ECSSD, DUT-O, PASCAL-S, HKU-IS, respectively. Our DFNet-L also improves the maxF and wF scores significantly. The results further demonstrate the effectiveness of our method in saliency detection task. It is worth noting that our method is end-to-end and does not need any post-processing methods such as CRF~\cite{krahenbuhl2011efficient}. Furthermore, our DFNet-V, DFNet-R, DFNet-M, and DFNet-L can
run at a speed of 32 FPS, 22 FPS, 26 FPS, and 9 FPS, respectively when processing a $352\times352$ image. One thing to note is that although our DFNet-M contains fewer parameters than all the other methods, it has great
performance, and it also can run at a real-time speed.

\begin{figure}[!tp]
\begin{center}
\includegraphics[width=1\textwidth]{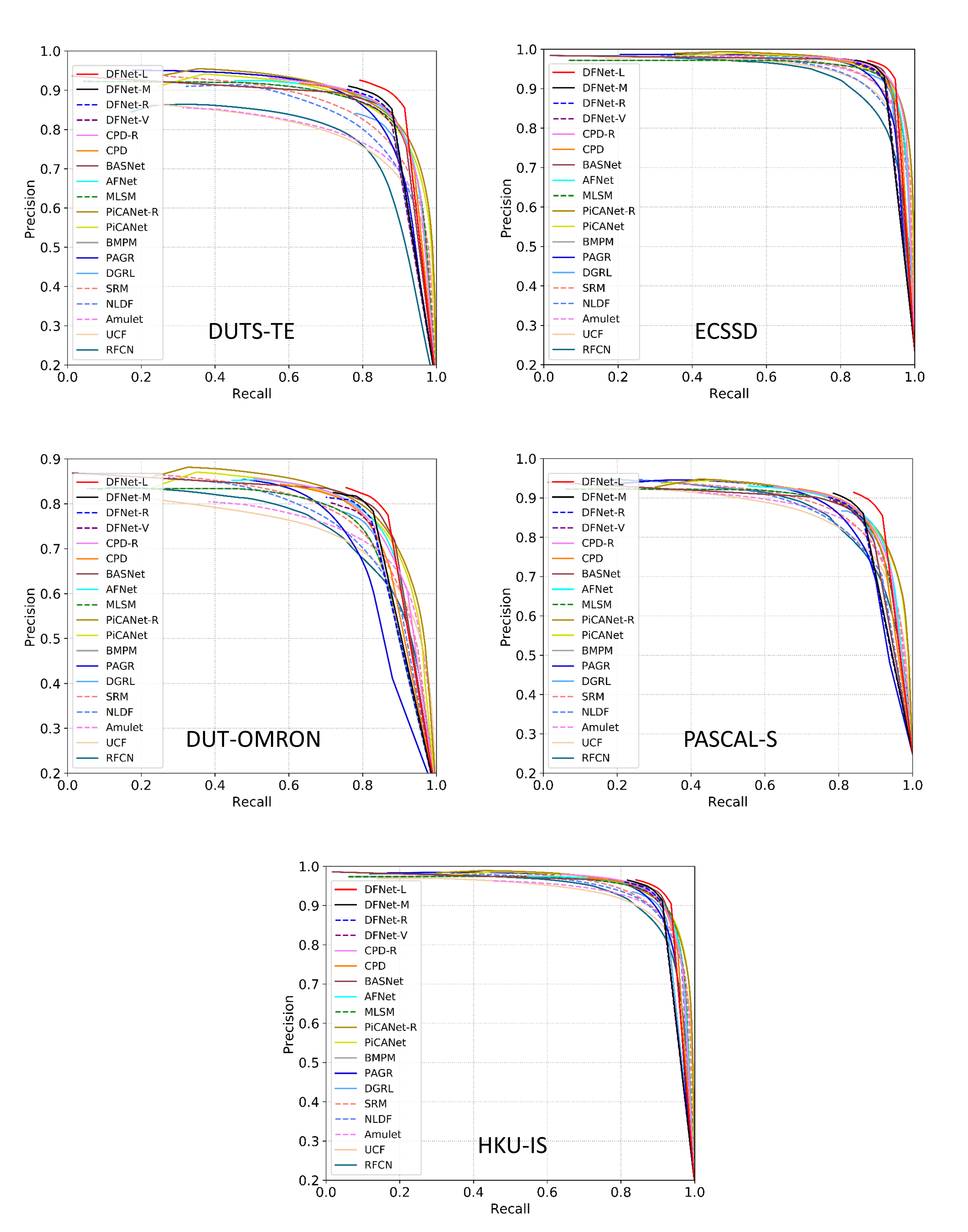}
\end{center}
\caption{
PR curves of the proposed method and previous state-of-the-art methods on five datasets.}
\label{fig:PRcurve}
\end{figure}

\begin{figure}[!tp]
\begin{center}
\includegraphics[width=1\textwidth]{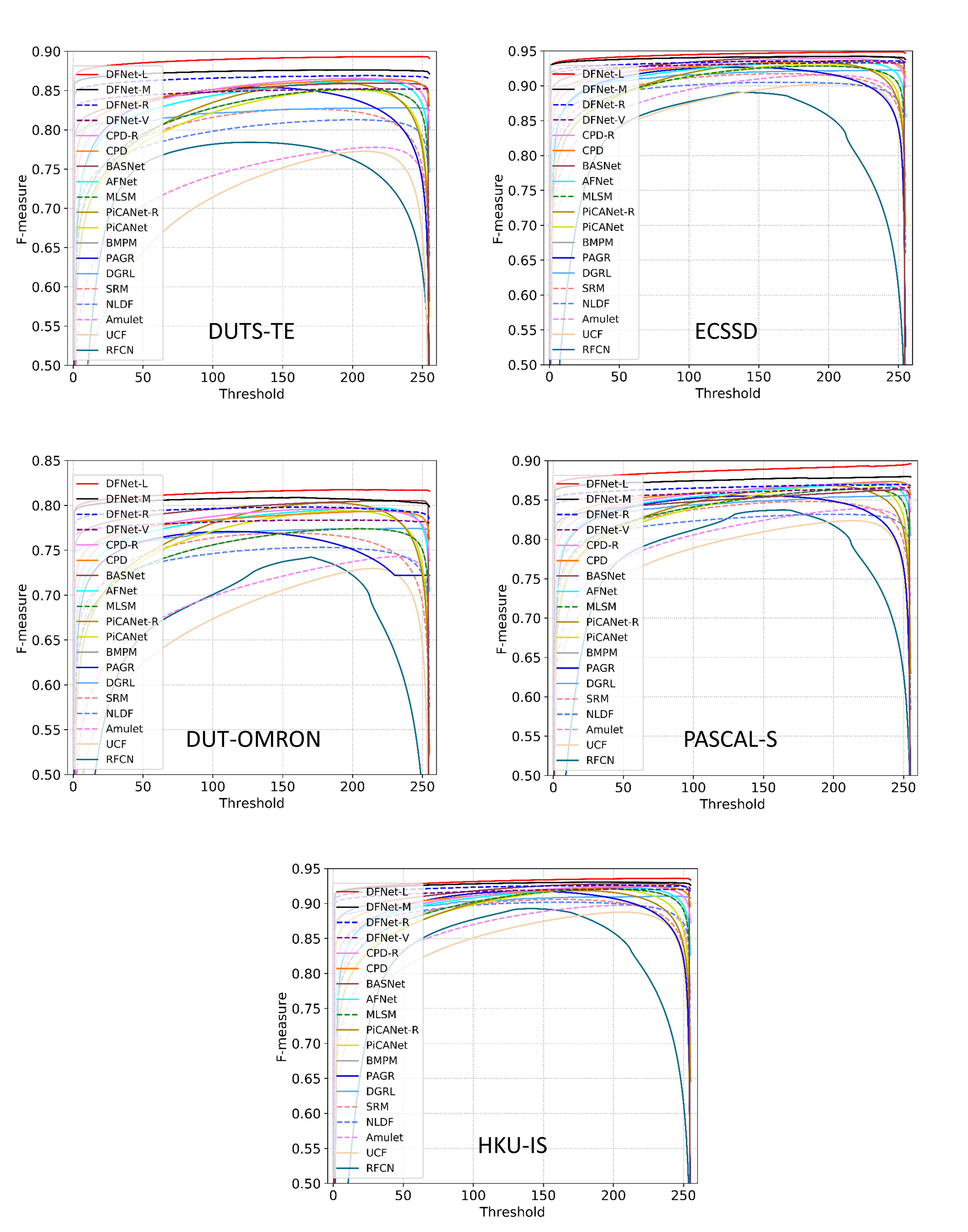}
\end{center}
\caption{F-measure curves of the proposed method and previous state-of-the-art methods on five datasets.}
\label{fig:Fcurve}
\end{figure}

\begin{table}[!tp]
\caption{The avgF, wF, maxF, and MAE scores of different saliency detection methods on five datasets. The best score under each setting is shown in {\color[HTML]{FE0000}{\textbf{red}}}, the second best score under each setting is shown in {\color[HTML]{3166FF}{\textbf{blue}}}, and the best score under all settings is underlined. DFNet with VGG-16, ResNet50, NASNet-Mobile, and NASNet-Large backbones, are denoted as DFNet-V, DFNet-R, DFNet-M, and DFNet-L, respectively. The unit of the total number of parameters (denoted as \#Par) is million. Note that the authors of \cite{zhang2018progressive} did not release the code, and they just provided the saliency maps, and thus reporting the total number of parameters is not possible for this method.}
\begin{center}
\resizebox{\textwidth}{!}{%
{\renewcommand{\arraystretch}{1.6}
\setlength\tabcolsep{2pt}
\begin{tabular}{lccccccccccccccccccccc}
\midrule
\multicolumn{1}{l|}{\textbf{Dataset}}   & \multicolumn{1}{c|}{}                                            & \multicolumn{4}{c|}{DUTS-TE~\cite{wang2017learning}}                                                                                                                                                                               & \multicolumn{4}{c|}{ECSSD~\cite{yang2013saliency}}                                                                                                                                                                                 & \multicolumn{4}{c|}{DUT-O~\cite{yang2013saliency}}                                                                                                                                                                                 & \multicolumn{4}{c|}{PASCAL-S~\cite{li2014secrets}}                                                                                                                                                                              & \multicolumn{4}{c}{HKU-IS~\cite{li2015visual}}                                                                                                                                                           \\ \cline{1-1} \cline{3-22} 
\multicolumn{1}{l|}{\textbf{Metric}}    & \multicolumn{1}{c|}{\multirow{-2}{*}{\#Par}}                     & avgF                                        & wF                                          & maxF                                        & \multicolumn{1}{c|}{MAE}                                         & avgF                                        & wF                                          & maxF                                        & \multicolumn{1}{c|}{MAE}                                         & avgF                                        & wF                                          & maxF                                        & \multicolumn{1}{c|}{MAE}                                         & avgF                                        & wF                                          & maxF                                        & \multicolumn{1}{c|}{MAE}                                         & avgF                                        & wF                                          & maxF                                        & MAE                                         \\ \midrule

\multicolumn{22}{c}{VGG16~\cite{simonyan2014very} backbone}                                                                                                                                                                                                                                                                                                                                                                                                                                                                                                                                                                                                                                                                                                                                                                                                                                                                                                                                                                                                                                                                                                              \\ \midrule
\multicolumn{1}{l|}{\textbf{MDF}~\cite{li2015visual}}       & \multicolumn{1}{c|}{56.86}                                       & 0.669                                       & 0.588                                       & 0.729                                       & \multicolumn{1}{c|}{0.093}                                       & 0.807                                       & 0.705                                       & 0.832                                       & \multicolumn{1}{c|}{0.105}                                       & 0644                                        & 0.564                                       & 0.694                                       & \multicolumn{1}{c|}{0.092}                                       & 0.711                                       & 0.590                                       & 0.770                                       & \multicolumn{1}{c|}{0.146}                                       & 0.784                                       & 0.564                                       & 0.860                                       & 0.129                                       \\
\multicolumn{1}{l|}{\textbf{RFCN}~\cite{wang2016saliency}}      & \multicolumn{1}{c|}{134.69}                                      & 0.711                                       & 0.586                                       & 0.784                                       & \multicolumn{1}{c|}{0.090}                                       & 0.834                                       & 0.698                                       & 0.890                                       & \multicolumn{1}{c|}{0.107}                                       & 0.627                                       & 0.524                                       & 0.742                                       & \multicolumn{1}{c|}{0.110}                                       & 0.754                                       & 0.636                                       & 0.837                                       & \multicolumn{1}{c|}{0.132}                                       & 0.835                                       & 0.680                                       & 0.893                                       & 0.089                                       \\
\multicolumn{1}{l|}{\textbf{DHS}~\cite{liu2016dhsnet}}       & \multicolumn{1}{c|}{94.04}                                       & 0.720                                       & 0.698                                       & 0.808                                       & \multicolumn{1}{c|}{0.067}                                       & 0.872                                       & 0.840                                       & 0.906                                       & \multicolumn{1}{c|}{0.059}                                       & ---                                         & ---                                         & ---                                         & \multicolumn{1}{c|}{---}                                         & 0.780                                       & 0.737                                       & 0.832                                       & \multicolumn{1}{c|}{0.094}                                       & 0.855                                       & 0.815                                       & 0.890                                       & 0.052                                       \\
\multicolumn{1}{l|}{\textbf{UCF}~\cite{zhang2017learning}}       & \multicolumn{1}{c|}{23.98}                                       & 0.631                                       & 0.596                                       & 0.773                                       & \multicolumn{1}{c|}{0.112}                                       & 0.844                                       & 0.806                                       & 0.903                                       & \multicolumn{1}{c|}{0.069}                                       & 0.621                                       & 0.573                                       & 0.730                                       & \multicolumn{1}{c|}{0.120}                                       & 0.738                                       & 0.700                                       & 0.824                                       & \multicolumn{1}{c|}{0.116}                                       & 0.823                                       & 0.779                                       & 0.888                                       & 0.062                                       \\
\multicolumn{1}{l|}{\textbf{Amulet}~\cite{zhang2017amulet}}    & \multicolumn{1}{c|}{33.15}                                       & 0.678                                       & 0.658                                       & 0.778                                       & \multicolumn{1}{c|}{0.085}                                       & 0.868                                       & 0.840                                       & 0.915                                       & \multicolumn{1}{c|}{0.059}                                       & 0.647                                       & 0.626                                       & 0.743                                       & \multicolumn{1}{c|}{0.098}                                       & 0.771                                       & 0.741                                       & 0.839                                       & \multicolumn{1}{c|}{0.099}                                       & 0.841                                       & 0.817                                       & 0.897                                       & 0.051                                       \\
\multicolumn{1}{l|}{\textbf{NLDF}~\cite{luo2017non}}      & \multicolumn{1}{c|}{35.49}                                       & 0.739                                       & 0.710                                       & 0.813                                       & \multicolumn{1}{c|}{0.065}                                       & 0.878                                       & 0.839                                       & 0.905                                       & \multicolumn{1}{c|}{0.063}                                       & 0.684                                       & 0.634                                       & 0.753                                       & \multicolumn{1}{c|}{0.080}                                       & 0.782                                       & 0.742                                       & 0.832                                       & \multicolumn{1}{c|}{0.101}                                       & 0.873                                       & 0.838                                       & 0.902                                       & 0.048                                       \\
\multicolumn{1}{l|}{\textbf{DSS}~\cite{hou2017deeply}}       & \multicolumn{1}{c|}{62.23}                                       & 0.716                                       & 0.702                                       & 0.813                                       & \multicolumn{1}{c|}{0.065}                                       & 0.873                                       & 0.836                                       & 0.908                                       & \multicolumn{1}{c|}{0.062}                                       & 0.674                                       & 0.643                                       & 0.760                                       & \multicolumn{1}{c|}{0.074}                                       & 0.776                                       & 0.728                                       & 0.832                                       & \multicolumn{1}{c|}{0.103}                                       & 0.856                                       & 0.821                                       & 0.900                                       & 0.050                                       \\
\multicolumn{1}{l|}{\textbf{RAS}~\cite{chen2018reverse}}       & \multicolumn{1}{c|}{{\color[HTML]{FE0000} \textbf{20.13}}}       & 0.751                                       & 0.740                                       & 0.831                                       & \multicolumn{1}{c|}{0.059}                                       & 0.889                                       & 0.857                                       & 0.921                                       & \multicolumn{1}{c|}{0.056}                                       & 0.713                                       & 0.695                                       & 0.786                                       & \multicolumn{1}{c|}{0.062}                                       & 0.787                                       & 0.738                                       & 0.836                                       & \multicolumn{1}{c|}{0.106}                                       & 0.871                                       & 0.843                                       & 0.913                                       & 0.045                                       \\
\multicolumn{1}{l|}{\textbf{BMPM}~\cite{zhang2018bi}}      & \multicolumn{1}{c|}{22.09}                                       & 0.745                                       & 0.761                                       & 0.851                                       & \multicolumn{1}{c|}{0.049}                                       & 0.868                                       & 0.871                                       & 0.928                                       & \multicolumn{1}{c|}{0.045}                                       & 0.692                                       & 0.681                                       & 0.774                                       & \multicolumn{1}{c|}{0.064}                                       & 0.771                                       & 0.785                                       & 0.864                                       & \multicolumn{1}{c|}{0.075}                                       & 0.871                                       & 0.859                                       & 0.921                                       & 0.039                                       \\
\multicolumn{1}{l|}{\textbf{PAGR}~\cite{zhang2018progressive}}      & \multicolumn{1}{c|}{---}                                         & 0.784                                       & 0.724                                       & 0.854                                       & \multicolumn{1}{c|}{0.055}                                       & 0.894                                       & 0.833                                       & 0.927                                       & \multicolumn{1}{c|}{0.061}                                       & 0.711                                       & 0.622                                       & 0.771                                       & \multicolumn{1}{c|}{0.071}                                       & 0.808                                       & 0.738                                       & 0.855                                       & \multicolumn{1}{c|}{0.095}                                       & 0.886                                       & 0.820                                       & 0.918                                       & 0.047                                       \\
\multicolumn{1}{l|}{\textbf{PiCANet}~\cite{liu2018picanet}}   & \multicolumn{1}{c|}{32.85}                                       & 0.749                                       & 0.747                                       & 0.851                                       & \multicolumn{1}{c|}{0.054}                                       & 0.885                                       & 0.865                                       & 0.931                                       & \multicolumn{1}{c|}{0.046}                                       & 0.710                                       & 0.691                                       & {\color[HTML]{3166FF} \textbf{0.794}}       & \multicolumn{1}{c|}{0.068}                                       & 0.804                                       & 0.781                                       & 0.870                                       & \multicolumn{1}{c|}{0.079}                                       & 0.870                                       & 0.847                                       & 0.921                                       & 0.042                                       \\
\multicolumn{1}{l|}{\textbf{MLMS}~\cite{wu2019mutual}}      & \multicolumn{1}{c|}{74.38}                                       & 0.745                                       & 0.761                                       & 0.852                                       & \multicolumn{1}{c|}{0.049}                                       & 0.868                                       & 0.871                                       & 0.928                                       & \multicolumn{1}{c|}{0.044}                                       & 0.692                                       & 0.681                                       & 0.774                                       & \multicolumn{1}{c|}{0.064}                                       & 0.771                                       & 0.785                                       & 0.864                                       & \multicolumn{1}{c|}{0.075}                                       & 0.871                                       & 0.859                                       & 0.921                                       & 0.039                                       \\
\multicolumn{1}{l|}{\textbf{AFNet}~\cite{feng2019attentive}}     & \multicolumn{1}{c|}{{\color[HTML]{3531FF} \textbf{21.08}}}       & 0.793                                       & 0.785                                       & {\color[HTML]{3166FF} \textbf{0.863}}       & \multicolumn{1}{c|}{0.046}                                       & 0.908                                       & 0.886                                       & {\color[HTML]{3166FF} \textbf{0.935}}       & \multicolumn{1}{c|}{0.042}                                       & 0.738                                       & {\color[HTML]{3166FF} \textbf{0.717}}       & {\color[HTML]{FE0000} \textbf{0.797}}       & \multicolumn{1}{c|}{{\color[HTML]{FE0000} \textbf{0.057}}}       & 0.828                                       & {\color[HTML]{3166FF} \textbf{0.804}}       & {\color[HTML]{3166FF} \textbf{0.871}}       & \multicolumn{1}{c|}{{\color[HTML]{FE0000} \textbf{0.071}}}       & 0.888                                       & 0.869                                       & {\color[HTML]{3166FF} \textbf{0.923}}       & 0.036                                       \\
\multicolumn{1}{l|}{\textbf{CPD}~\cite{wu2019cascaded}}       & \multicolumn{1}{c|}{29.32}                                       & {\color[HTML]{3166FF} \textbf{0.813}}       & {\color[HTML]{FE0000} \textbf{0.801}}       & {\color[HTML]{FE0000} \textbf{0.864}}       & \multicolumn{1}{c|}{{\color[HTML]{FE0000} \textbf{0.043}}}       & {\color[HTML]{3166FF} \textbf{0.914}}       & {\color[HTML]{3166FF} \textbf{0.895}}       & {\color[HTML]{FE0000} \textbf{0.936}}       & \multicolumn{1}{c|}{{\color[HTML]{FE0000} \textbf{0.040}}}       & {\color[HTML]{3166FF} \textbf{0.745}}       & 0.715                                       & {\color[HTML]{3166FF} \textbf{0.794}}       & \multicolumn{1}{c|}{{\color[HTML]{FE0000} \textbf{0.057}}}       & {\color[HTML]{3166FF} \textbf{0.832}}       & {\color[HTML]{FE0000} \textbf{0.806}}       & {\color[HTML]{FE0000} \textbf{0.873}}       & \multicolumn{1}{c|}{{\color[HTML]{3166FF} \textbf{0.074}}}       & {\color[HTML]{3166FF} \textbf{0.895}}       & {\color[HTML]{3166FF} \textbf{0.879}}       & {\color[HTML]{FE0000} \textbf{0.924}}       & {\color[HTML]{FE0000} \textbf{0.033}}       \\
\multicolumn{1}{l|}{\textbf{DFNet-V}}   & \multicolumn{1}{c|}{27.04}                                       & {\color[HTML]{FE0000} \textbf{0.824}}       & {\color[HTML]{3166FF} \textbf{0.796}}       & 0.852                                       & \multicolumn{1}{c|}{{\color[HTML]{3166FF} \textbf{0.045}}}       & {\color[HTML]{FE0000} \textbf{0.919}}       & {\color[HTML]{FE0000} \textbf{0.897}}       & 0.933                                       & \multicolumn{1}{c|}{{\color[HTML]{FE0000} \textbf{0.040}}}       & {\color[HTML]{FE0000} \textbf{0.751}}       & {\color[HTML]{FE0000} \textbf{0.721}}       & 0.784                                       & \multicolumn{1}{c|}{{\color[HTML]{FE0000} \textbf{0.057}}}       & {\color[HTML]{FE0000} \textbf{0.837}}       & 0.803                                       & 0.866                                       & \multicolumn{1}{c|}{0.075}                                       & {\color[HTML]{FE0000} \textbf{0.906}}       & {\color[HTML]{FE0000} \textbf{0.884}}       & 0.921                                       & {\color[HTML]{FE0000} \textbf{0.033}}       \\ \midrule
\multicolumn{22}{c}{ResNet50~\cite{he2016deep} backbone}                                                                                                                                                                                                                                                                                                                                                                                                                                                                                                                                                                                                                                                                                                                                                                                                                                                                                                                                                                                                                                                                                                          \\ \midrule
\multicolumn{1}{l|}{\textbf{SRM}~\cite{wang2017stagewise}}       & \multicolumn{1}{c|}{{\color[HTML]{3166FF} \textbf{43.74}}}       & 0.753                                       & 0.722                                       & 0.826                                       & \multicolumn{1}{c|}{0.059}                                       & 0.892                                       & 0.853                                       & 0.917                                       & \multicolumn{1}{c|}{0.054}                                       & 0.707                                       & 0.658                                       & 0.769                                       & \multicolumn{1}{c|}{0.069}                                       & 0.803                                       & 0.762                                       & 0.848                                       & \multicolumn{1}{c|}{0.087}                                       & 0.874                                       & 0.835                                       & 0.906                                       & 0.046                                       \\
\multicolumn{1}{l|}{\textbf{DGRL}~\cite{wang2018detect}}      & \multicolumn{1}{c|}{126.35}                                      & 0.794                                       & 0.774                                       & 0.828                                       & \multicolumn{1}{c|}{0.050}                                       & 0.906                                       & 0.891                                       & 0.922                                       & \multicolumn{1}{c|}{0.041}                                       & 0.733                                       & 0.709                                       & 0.774                                       & \multicolumn{1}{c|}{0.062}                                       & 0.827                                       & 0.802                                       & 0.856                                       & \multicolumn{1}{c|}{0.073}                                       & 0.890                                       & 0.875                                       & 0.910                                       & 0.036                                       \\
\multicolumn{1}{l|}{\textbf{PiCANet-R}~\cite{liu2018picanet}} & \multicolumn{1}{c|}{{\color[HTML]{FE0000} \textbf{37.02}}}       & 0.759                                       & 0.755                                       & 0.860                                       & \multicolumn{1}{c|}{0.051}                                       & 0.886                                       & 0.867                                       & 0.935                                       & \multicolumn{1}{c|}{0.046}                                       & 0.717                                       & 0.695                                       & 0.803                                       & \multicolumn{1}{c|}{0.065}                                       & 0.804                                       & 0.782                                       & 0.868                                       & \multicolumn{1}{c|}{0.078}                                       & 0.870                                       & 0.840                                       & 0.918                                       & 0.043                                       \\
\multicolumn{1}{l|}{\textbf{CapSal}~\cite{zhang2019capsal}}    & \multicolumn{1}{c|}{91.09}                                       & 0.755                                       & 0.689                                       & 0.819                                       & \multicolumn{1}{c|}{0.063}                                       & ---                                         & ---                                         & ---                                         & \multicolumn{1}{c|}{---}                                         & ---                                         & ---                                         & ---                                         & \multicolumn{1}{c|}{---}                                         & 0.827                                       & 0.791                                       & 0.869                                       & \multicolumn{1}{c|}{0.074}                                       & 0.841                                       & 0.780                                       & 0.880                                       & 0.058                                       \\
\multicolumn{1}{l|}{\textbf{BASNet}~\cite{qin2019basnet}}    & \multicolumn{1}{c|}{87.06}                                       & 0.791                                       & {\color[HTML]{3166FF} \textbf{0.803}}       & 0.860                                       & \multicolumn{1}{c|}{0.047}                                       & 0.880                                       & {\color[HTML]{FE0000} \textbf{0.904}}       & {\color[HTML]{FE0000} \textbf{0.942}}       & \multicolumn{1}{c|}{{\color[HTML]{FE0000} \textbf{0.037}}}       & {\color[HTML]{3166FF} \textbf{0.756}}       & {\color[HTML]{FE0000} \textbf{0.751}}       & {\color[HTML]{FE0000} \textbf{0.805}}       & \multicolumn{1}{c|}{{\color[HTML]{3166FF} \textbf{0.056}}}       & 0.781                                       & 0.800                                       & 0.863                                       & \multicolumn{1}{c|}{0.077}                                       & {\color[HTML]{3166FF} \textbf{0.895}}       & {\color[HTML]{FE0000} \textbf{0.889}}       & {\color[HTML]{FE0000} \textbf{0.928}}       & {\color[HTML]{FE0000} \textbf{0.032}}       \\
\multicolumn{1}{l|}{\textbf{CPD-R}~\cite{wu2019cascaded}}     & \multicolumn{1}{c|}{47.85}                                       & {\color[HTML]{3166FF} \textbf{0.805}}       & 0.795                                       & {\color[HTML]{3166FF} \textbf{0.865}}       & \multicolumn{1}{c|}{{\color[HTML]{3166FF} \textbf{0.043}}}       & {\color[HTML]{3166FF} \textbf{0.917}}       & 0.898                                       & {\color[HTML]{3166FF} \textbf{0.939}}       & \multicolumn{1}{c|}{{\color[HTML]{FE0000} \textbf{0.037}}}       & 0.747                                       & 0.719                                       & 0.797                                       & \multicolumn{1}{c|}{{\color[HTML]{3166FF} \textbf{0.056}}}       & {\color[HTML]{3166FF} \textbf{0.831}}       & {\color[HTML]{3166FF} \textbf{0.803}}       & {\color[HTML]{FE0000} \textbf{0.872}}       & \multicolumn{1}{c|}{{\color[HTML]{3166FF} \textbf{0.072}}}       & 0.891                                       & 0.875                                       & 0.925                                       & 0.034                                       \\
\multicolumn{1}{l|}{\textbf{DFNet-R}}   & \multicolumn{1}{c|}{54.74}                                       & {\color[HTML]{FE0000} \textbf{0.845}}       & {\color[HTML]{FE0000} \textbf{0.817}}       & {\color[HTML]{FE0000} \textbf{0.869}}       & \multicolumn{1}{c|}{{\color[HTML]{FE0000} \textbf{0.040}}}       & {\color[HTML]{FE0000} \textbf{0.922}}       & {\color[HTML]{3166FF} \textbf{0.899}}       & 0.936                                       & \multicolumn{1}{c|}{0.039}                                       & {\color[HTML]{FE0000} \textbf{0.766}}       & {\color[HTML]{3166FF} \textbf{0.734}}       & {\color[HTML]{3166FF} \textbf{0.798}}       & \multicolumn{1}{c|}{{\color[HTML]{FE0000} \textbf{0.053}}}       & {\color[HTML]{FE0000} \textbf{0.845}}       & {\color[HTML]{FE0000} \textbf{0.811}}       & {\color[HTML]{3166FF} \textbf{0.870}}       & \multicolumn{1}{c|}{{\color[HTML]{FE0000} \textbf{0.070}}}       & {\color[HTML]{FE0000} \textbf{0.912}}       & {\color[HTML]{FE0000} \textbf{0.889}}       & {\color[HTML]{3166FF} \textbf{0.926}}       & {\color[HTML]{FE0000} \textbf{0.032}}       \\ \midrule
\multicolumn{22}{c}{NASNet~\cite{zoph2018learning} backbone}                                                                                                                                                                                                                                                                                                                                                                                                                                                                                                                                                                                                                                                                                                                                                                                                                                                                                                                                                                                                                                                                                                      \\  \midrule
\multicolumn{1}{l|}{\textbf{DFNet-M}}   & \multicolumn{1}{c|}{{\color[HTML]{FE0000} \textbf{\underline{19.02}}}} & {\color[HTML]{3166FF} \textbf{0.855}}       & {\color[HTML]{3166FF} \textbf{0.827}}       & {\color[HTML]{3166FF} \textbf{0.876}}       & \multicolumn{1}{c|}{{\color[HTML]{3166FF} \textbf{0.038}}}       & {\color[HTML]{3166FF} \textbf{0.930}}       & {\color[HTML]{3166FF} \textbf{0.908}}       & {\color[HTML]{3166FF} \textbf{0.942}}       & \multicolumn{1}{c|}{{\color[HTML]{3166FF} \textbf{0.037}}}       & {\color[HTML]{3166FF} \textbf{0.777}}       & {\color[HTML]{3166FF} \textbf{0.748}}       & {\color[HTML]{3166FF} \textbf{0.809}}       & \multicolumn{1}{c|}{{\color[HTML]{FE0000} \textbf{\underline{0.052}}}} & {\color[HTML]{3166FF} \textbf{0.854}}       & {\color[HTML]{3166FF} \textbf{0.821}}       & {\color[HTML]{3166FF} \textbf{0.880}}       & \multicolumn{1}{c|}{{\color[HTML]{3166FF} \textbf{0.068}}}       & {\color[HTML]{3166FF} \textbf{0.918}}       & {\color[HTML]{3166FF} \textbf{0.897}}       & {\color[HTML]{3166FF} \textbf{0.931}}       & {\color[HTML]{3166FF} \textbf{0.030}}       \\
\multicolumn{1}{l|}{\textbf{DFNet-L}}   & \multicolumn{1}{c|}{{\color[HTML]{3166FF} \textbf{127.55}}}      & {\color[HTML]{FE0000} \textbf{\underline{0.873}}} & {\color[HTML]{FE0000} \textbf{\underline{0.854}}} & {\color[HTML]{FE0000} \textbf{\underline{0.893}}} & \multicolumn{1}{c|}{{\color[HTML]{FE0000} \textbf{\underline{0.033}}}} & {\color[HTML]{FE0000} \textbf{\underline{0.937}}} & {\color[HTML]{FE0000} \textbf{\underline{0.923}}} & {\color[HTML]{FE0000} \textbf{\underline{0.949}}} & \multicolumn{1}{c|}{{\color[HTML]{FE0000} \textbf{\underline{0.028}}}} & {\color[HTML]{FE0000} \textbf{\underline{0.789}}} & {\color[HTML]{FE0000} \textbf{\underline{0.769}}} & {\color[HTML]{FE0000} \textbf{\underline{0.817}}} & \multicolumn{1}{c|}{{\color[HTML]{FE0000} \textbf{\underline{0.052}}}} & {\color[HTML]{FE0000} \textbf{\underline{0.873}}} & {\color[HTML]{FE0000} \textbf{\underline{0.854}}} & {\color[HTML]{FE0000} \textbf{\underline{0.896}}} & \multicolumn{1}{c|}{{\color[HTML]{FE0000} \textbf{\underline{0.054}}}} & {\color[HTML]{FE0000} \textbf{\underline{0.923}}} & {\color[HTML]{FE0000} \textbf{\underline{0.908}}} & {\color[HTML]{FE0000} \textbf{\underline{0.936}}} & {\color[HTML]{FE0000} \textbf{\underline {0.028}}} \\ \midrule
\end{tabular}
}
}
\end{center}
\label{tab:comparison}
\end{table}

\begin{figure}[!tp]
\centering
\includegraphics[width=1\textwidth]{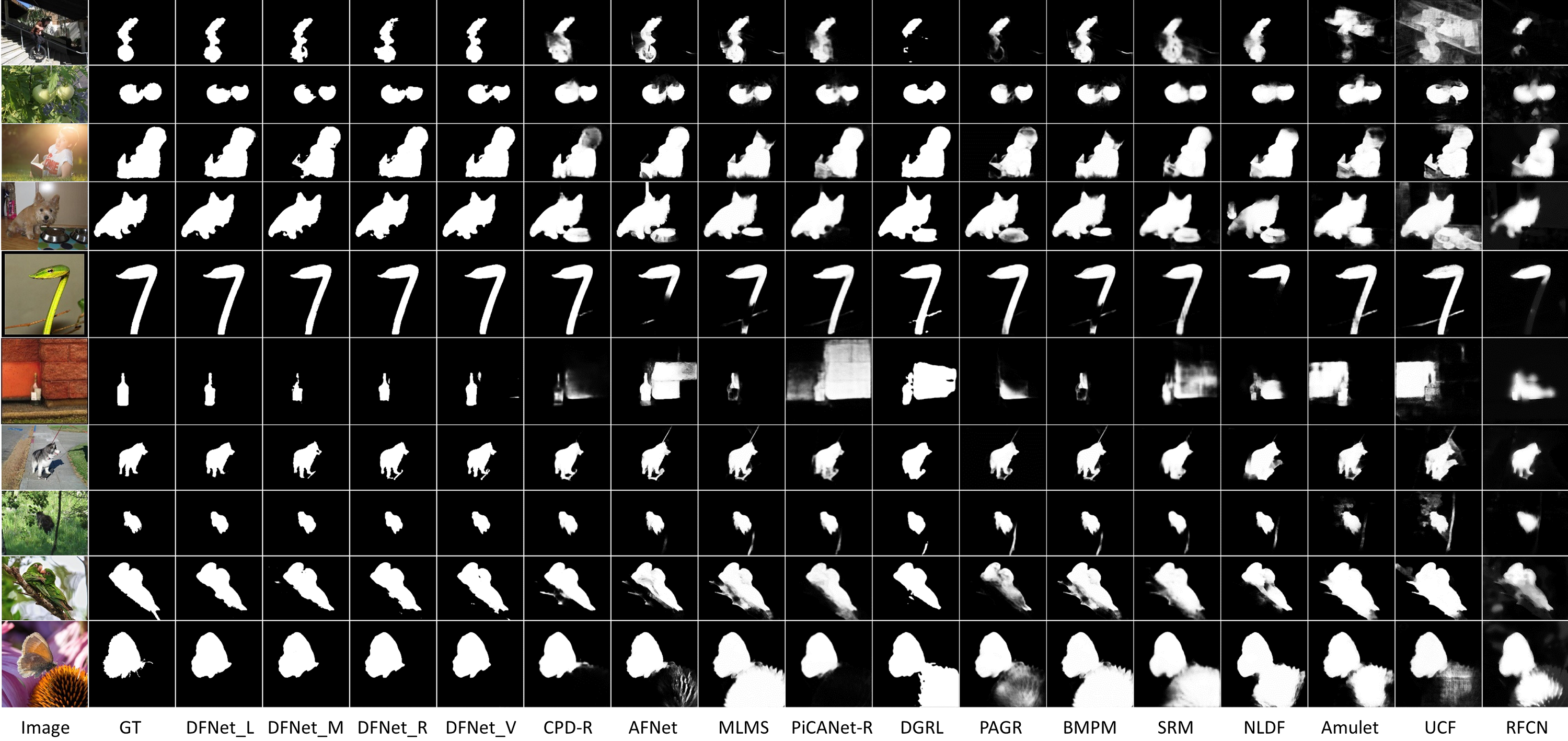}
\caption{Qualitative comparison between our results and state-of-the-art methods. Our model outputs sharper saliency maps that are closer to the ground truth masks compared to other models.}
\label{fig:visualexamples}
\end{figure}

For qualitative evaluation, we show a visual comparison between our method and previous state-of-the-art salient object detection methods in Figure \ref{fig:visualexamples}. It can be seen that our approach can uniformly highlight the inner part of salient regions in various challenging and complex scenes. Our model is also capable of suppressing the background regions that are wrongly predicted as salient by other methods. As seen from Figure \ref{fig:visualexamples}, by taking advantage of the adopted modules in our framework and the Sharpening Loss, our model predicts saliency maps that are closer to the ground truth masks compared to other methods.

\subsection{Ablation Study}

In this section, we conduct experiments on DFNet-V to investigate the effectiveness of different components in our method. The results are provided in Table \ref{tab:abl}. The details of these experiments are explained below.
\subsubsection{The effectiveness of MAG Modules}

 To show the effectiveness of MAG Modules, we remove them from the network, which is denoted as Without MAG in Table \ref{tab:abl}. As seen in this table, the performance degrades over all datasets and evaluation metrics. The results confirm that the proposed module is helpful for salient object detection.

\subsubsection{The effectiveness of AMI Modules}

We demonstrate the effectiveness of AMI Modules by replacing them with Concatenation layers, which is denoted as Without AMI in Table \ref{tab:abl}. From this table, we can see that the performance gets worse.
Additionally, we remove the MAG Modules and also replace AMI Modules with Concatenation layers, which is denoted as Without MAG and AMI in Table \ref{tab:abl}. As seen from this table, the performance gets worse drastically. The results prove the beneficial effect of our modules in salient object detection.

\subsubsection{The effectiveness of CA Blocks}
 As previously explained, we use CA Blocks in the MAG and AMI Modules. To demonstrate their effectiveness in our network, we remove them, which is denoted as Without CAs in Table \ref{tab:abl}. From this table, we can see that the performance degrades, which shows that using CA Blocks have beneficial effects on the final results.

\subsubsection{The effectiveness of the Sharpening Loss function}
 To validate our choice of loss function (Section \ref{ch:learning_sharper}), we train DFNet-V with Cross-entropy loss (denoted as Cross-entropy in Table \ref{tab:abl}) and compare it with the Sharpening Loss. Quantitative comparison in Table \ref{tab:abl} demonstrate that the proposed Sharpening Loss outperforms the widely-used Cross-entropy loss by a significant margin. For qualitative evaluation, a visual comparison between the Sharpening Loss and Cross-entropy loss is shown in Figure \ref{fig:ablation_sm}. As seen from this figure, our network trained with the Sharpening Loss, learns sharper salient objects compared to the one with the Cross-entropy loss. Thus, the Sharpening Loss guides our network to output saliency maps with higher certainty and less blurry salient objects which are much close to the ground truth compared to the Cross-entropy Loss.

In order to investigate the effect of the balance parameter $\lambda$ on the final results, we conduct experiments for different balance value. We test nine values from 0.5 to 2.5 with steps of 0.25. The results for MAE metric on all datasets are shown in Figure \ref{fig:lambda}. As shown in this figure, with the same setting, our method achieves the best performance with the value of 1.75, which means that choosing this value for $\lambda$ results in the best balance between the F-measure Loss $L_F$ and the MAE Loss $L_{MAE}$.

\begin{table}[!tp]
\begin{center}
\caption{Ablation analysis. The performance of different settings of our model (The best score is shown in {\color[HTML]{FE0000} \textbf{red}}).}
\label{tab:abl}
\resizebox{\textwidth}{!}{%
{\renewcommand{\arraystretch}{1.6}
\setlength\tabcolsep{2pt}
\begin{tabular}{l|cccc|cccc|cccc|cccc|cccc}
\midrule
\textbf{Dataset}                  & \multicolumn{4}{c|}{DUTS-TE~\cite{wang2017learning}}                                                                                                                                  & \multicolumn{4}{c|}{ECSSD~\cite{yang2013saliency}}                                                                                                                                    & \multicolumn{4}{c|}{DUT-O~\cite{yang2013saliency}}                                                                                                                                    & \multicolumn{4}{c|}{PASCAL-S~\cite{li2014secrets}}                                                                                                                                 & \multicolumn{4}{c}{HKU-IS~\cite{li2015visual}}                                                                                                                                   \\ \hline
\textbf{Metric}                   & avgF                                  & wF                                    & maxF                                  & MAE                                   & avgF                                  & wF                                    & maxF                                  & MAE                                   & avgF                                  & wF                                    & maxF                                  & MAE                                   & avgF                                  & wF                                    & maxF                                  & MAE                                   & avgF                                  & wF                                    & maxF                                   & MAE                                   \\ \midrule
\textbf{Without $MAG$}               & 0.810                                 & 0.767                                 & 0.841                                 & 0.049                                 & 0.900                                 & 0.864                                 & 0.920                                 & 0.052                                 & 0.728                                 & 0.682                                 & 0.768                                 & 0.063                                 & 0.815                                 & 0.767                                 & 0.850                                 & 0.089                                 & 0.894                                 & 0.860                                 & 0.912                                 & 0.039                                 \\
\textbf{Without $AMI$}              & 0.802                                 & 0.768                                 & 0.831                                 & 0.049                                 & 0.905                                 & 0.880                                 & 0.921                                 & 0.045                                 & 0.727                                 & 0.689                                 & 0.765                                 & 0.061                                 & 0.821                                 & 0.785                                 & 0.851                                 & 0.080                                 & 0.892                                 & 0.866                                 & 0.908                                 & 0.037                                 \\
\textbf{Without $MAG$ and $AMI$} & 0.737                                 & 0.697                                 & 0.768                                 & 0.065                                 & 0.864                                 & 0.827                                 & 0.885                                 & 0.065                                 & 0646                                  & 0.598                                 & 0.692                                 & 0.086                                 & 0.777                                 & 0.731                                 & 0.809                                 & 0.099                                 & 0.854                                 & 0.823                                 & 0.873                                 & 0.049                                 \\
\textbf{Without $CAs$}             & 0.808                                 & 0.778                                 & 0.838                                 & 0.048                                 & 0.913                                 & 0.889                                 & 0.929                                 & 0.043                                 & 0.735                                 & 0.703                                 & 0.771                                 & 0.060                                 & 0.828                                 & 0.794                                 & 0.859                                 & 0.077                                 & 0.899                                 & 0.875                                 & 0.915                                 & 0.035                                 \\
\textbf{Cross-entropy}            & 0.747                                 & 0.746                                 & 0.850                                 & 0.052                                 & 0.878                                 & 0.853                                 & 0.922                                 & 0.052                                 & 0.700                                 & 0.680                                 & 0.782                                 & 0.063                                 & 0.796                                 & 0.764                                 & 0.856                                 & 0.085                                 & 0.863                                 & 0.843                                 & 0.915                                 & 0.042                                 \\
\textbf{DFNet-V}         & {\color[HTML]{FE0000} \textbf{0.824}} & {\color[HTML]{FE0000} \textbf{0.796}} & {\color[HTML]{FE0000} \textbf{0.852}} & {\color[HTML]{FE0000} \textbf{0.045}} & {\color[HTML]{FE0000} \textbf{0.919}} & {\color[HTML]{FE0000} \textbf{0.897}} & {\color[HTML]{FE0000} \textbf{0.933}} & {\color[HTML]{FE0000} \textbf{0.040}} & {\color[HTML]{FE0000} \textbf{0.751}} & {\color[HTML]{FE0000} \textbf{0.721}} & {\color[HTML]{FE0000} \textbf{0.784}} & {\color[HTML]{FE0000} \textbf{0.057}} & {\color[HTML]{FE0000} \textbf{0.837}} & {\color[HTML]{FE0000} \textbf{0.803}} & {\color[HTML]{FE0000} \textbf{0.866}} & {\color[HTML]{FE0000} \textbf{0.075}} & {\color[HTML]{FE0000} \textbf{0.906}} & {\color[HTML]{FE0000} \textbf{0.884}} & {\color[HTML]{FE0000} \textbf{0.921}} & {\color[HTML]{FE0000} \textbf{0.033}} \\ \midrule
\end{tabular}
}
}
\end{center}
\end{table}

\begin{figure}[!bp]
\begin{center}
\includegraphics[width=0.55\textwidth]{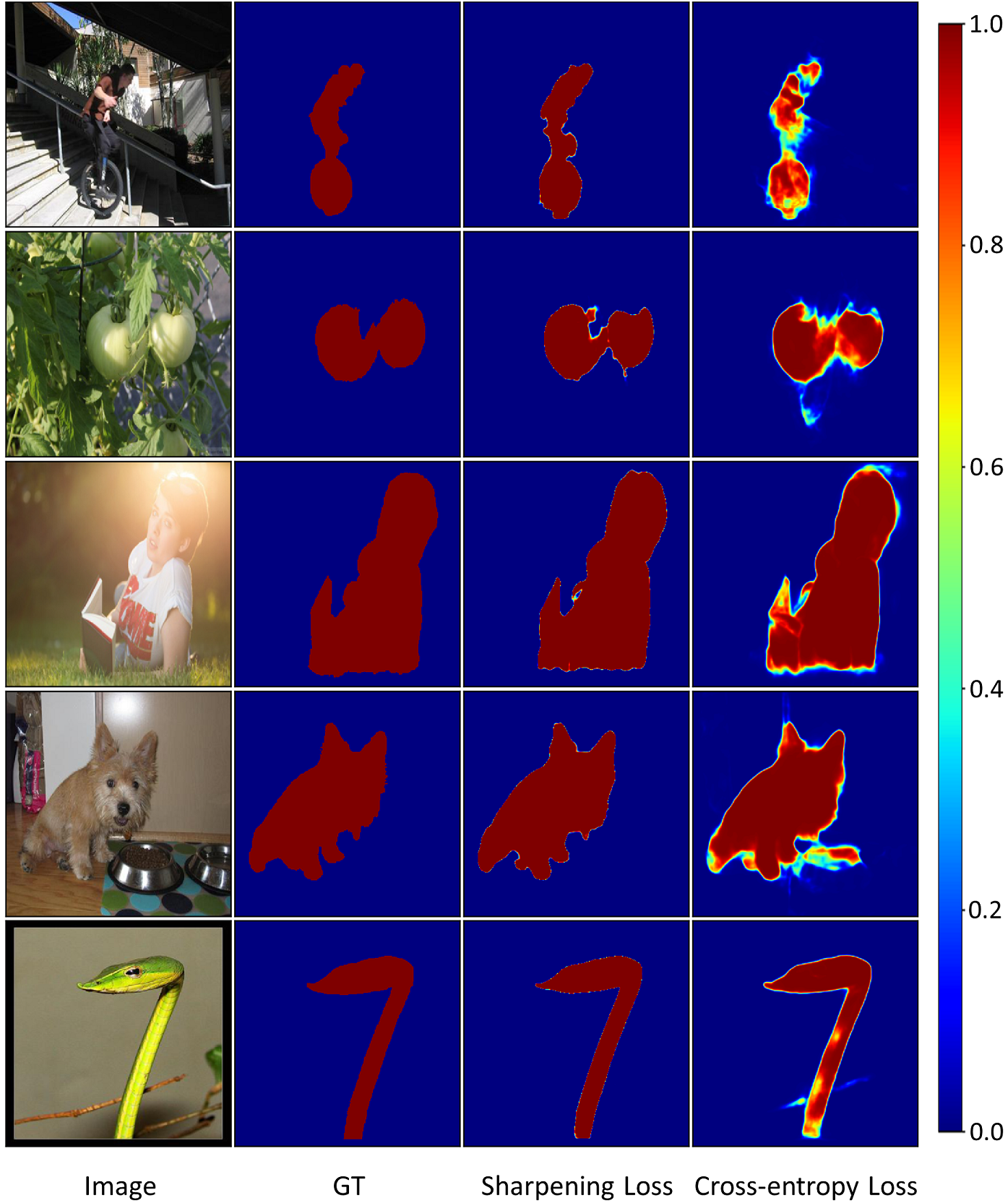}
\end{center}
\caption{Visual comparison between the Sharpening Loss and the Cross-entropy Loss. Our network trained with the Sharpening Loss learns sharper salient objects whose predicted values are much closer to 1. In contrast, our network trained with the Cross-entropy Loss outputs blurry saliency maps.}
\label{fig:ablation_sm}
\end{figure}

\begin{figure}[!tp]
\begin{center}
\includegraphics[width=0.55\textwidth]{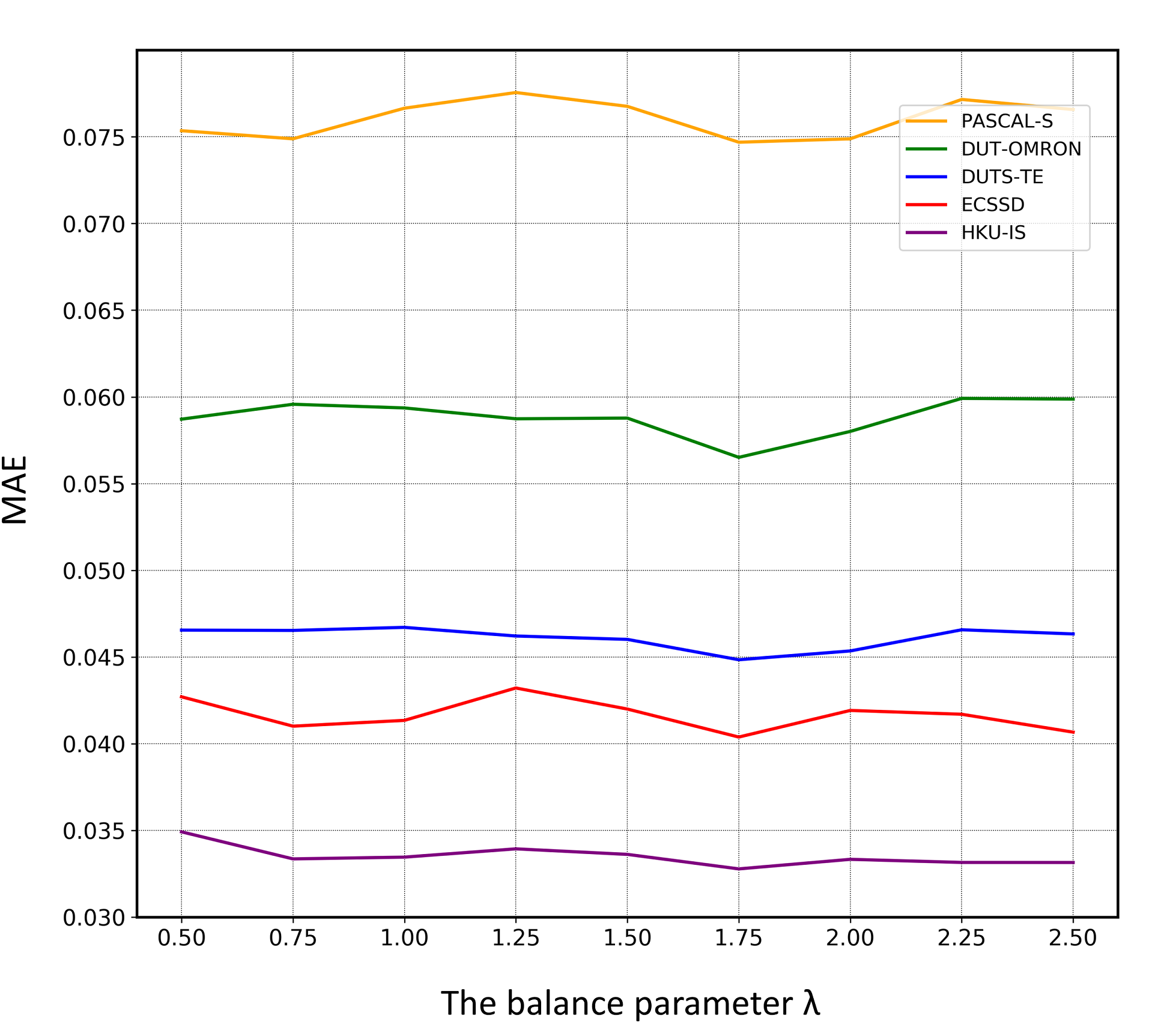}
\end{center}
\caption{Results of DFNet-V for MAE metric with different $\lambda$ value on all datasets.}
\label{fig:lambda}
\end{figure}

\section{Conclusion}
In this work, we introduce a robust and flexible framework for saliency detection task, which is composed of two main modules. The first one is the Multi-scale Attention Guided Module which extracts multi-scale features effectively, and then adaptively weights feature maps of various scales. By adopting this module, the model learns to give more attention to more discriminative feature maps corresponding to the scale of the salient object in the input image. The second module is the Attention-based Multi-level Integrator Module which gives the model the flexibility to assign different weights to multi-level feature maps. In addition, our Sharpening Loss function outperforms the Cross-entropy loss and leads to sharper salient objects. The proposed method achieves the state-of-the-art performance on several challenging datasets.


\section*{References}

\bibliography{DFNet}

\end{document}